\documentclass{article}
\usepackage[a4paper, margin=1.25in]{geometry} 

\usepackage{graphicx} 
\usepackage{url}
\usepackage{multirow} 
\usepackage{authblk}

\usepackage{array}
\usepackage{graphicx}
\usepackage{xspace}
\usepackage{enumerate}
\usepackage{courier} 
\usepackage[table,xcdraw]{xcolor}
\usepackage{amsmath}
\usepackage{amssymb}
\usepackage{algorithm2e}

\usepackage{multirow} 

\usepackage{xcolor}

\newcommand{\ie}{\textit{i}.\textit{e}.,\xspace}
\newcommand{\eg}{\textit{e}.\textit{g}.,\xspace}

\newcommand{\ontoOne}{\ensuremath{\On_1}}
\newcommand{\ontoTwo}{\ensuremath{\On_2}}

\newcommand{\logmap}{\texttt{LogMap}\xspace}
\newcommand{\logmapllm}{\texttt{LogMapLLM}\xspace}
\newcommand{\logmapP}{\texttt{LogMap}+}

\newcommand{\oracle}{Oracle\xspace}
\newcommand{\oracles}{Oracles\xspace}
\newcommand{\Oracle}{Oracle\xspace}
\newcommand{\Oracles}{Oracles\xspace}

\newcommand{\OrLLM}{\texttt{Or}$^{LLM}$\xspace}
\newcommand{\OrGem}{\texttt{Or}$^{LLM}_{GF2.0}$\xspace}
\newcommand{\OrGemNew}{\texttt{Or}$^{LLM}_{GF2.5}$\xspace}
\newcommand{\OrZero}{\texttt{Or}$^{0}$\xspace}
\newcommand{\OrTen}{\texttt{Or}$^{10}$\xspace}
\newcommand{\OrTwenty}{\texttt{Or}$^{20}$\xspace}
\newcommand{\OrThirty}{\texttt{Or}$^{30}$\xspace}

\newcommand{\On}{\ensuremath{\mathcal{O}}\xspace}
\newcommand{\M}{\ensuremath{\mathcal{M}}}

\newcommand{\mapping}[4]{\langle #1,\allowbreak #2,\allowbreak #3,\allowbreak #4
\rangle}
\newcommand{\mappingTwo}[2]{\langle #1,\allowbreak #2 \rangle}
\newcommand{\myset}[1]{ \{#1\} }
\newcommand{\MS}{\M_{S}\xspace}

\newcommand{\MRA}{\M_{RA}\xspace}

\newcommand{\Mask}{\M_{ask}\xspace}

\newcommand{\logicalalgo}[1]{\ifmmode \text{ \textbf{#1} } \else \textbf{#1} \fi}

\newcommand{\Prompt}{\ensuremath{\mathbf{P}}\xspace}

\newcommand{\PEC}{\ensuremath{\Prompt_{\mathrm{EC}}}\xspace}   
\newcommand{\PNLF}{\ensuremath{\Prompt^{\mathrm{NLF}}}\xspace} 

\newcommand{\PECNLF}{\ensuremath{\Prompt_{\mathrm{EC}}^{\mathrm{NLF}}}\xspace}
\newcommand{\PECSNLF}{\ensuremath{\Prompt_{\mathrm{EC+S}}^{\mathrm{NLF}}}\xspace}
\newcommand{\PSNLF}{\ensuremath{\Prompt_{\mathrm{S}}^{\mathrm{NLF}}}\xspace}

\newcommand{\change}[1]{{\color{black}#1}} %

\usepackage{hyperref}
\usepackage{color}

\SetKwInput{KwInput}{Input}
\SetKwInput{KwOutput}{Output}

\usepackage{subcaption}
\usepackage{float}

\usepackage{listings,xcolor}
\AtBeginDocument{%
  
}

\title{Large Language Models as Oracles for Ontology Alignment}
\author[1\dag]{Sviatoslav Lushnei}
\author[1\dag]{Dmytro Shumskyi}
\author[1\dag]{Severyn Shykula}
\author[2]{\\Ernesto Jimen\'ez-Ruiz\thanks{Corresponding author: \url{ernesto.jimenez-ruiz@citystgeorges.ac.uk}}}
\author[2]{Artur d'Avila Garcez}
\affil[1]{Ukrainian Catholic University, Ukraine}
\affil[2]{City St George's, University of London, UK}
\date{\normalsize Paper accepted at the 19th Conference of the European Chapter of the\\ Association for Computational Linguistics (EACL 2026), main conference.\\ \textsuperscript{\textcopyright} 2026 Association for Computational Linguists.}

\begin{document}

\def\thefootnote{\dag}\footnotetext{These authors contributed equally to this work}\def\thefootnote{\arabic{footnote}}

\maketitle

\begin{abstract}
There are many methods and systems to tackle the ontology alignment problem, yet a major challenge persists in producing high-quality mappings among a set of input ontologies. Adopting a human-in-the-loop approach during the alignment process has become essential in applications requiring very accurate mappings. However, user involvement is expensive when dealing with large ontologies. In this paper, we analyse the feasibility of using Large Language Models (LLM) to aid the ontology alignment problem. LLMs are used only in the validation of a subset of correspondences for which there is high uncertainty. We have conducted an extensive analysis over several tasks of the Ontology Alignment Evaluation Initiative (OAEI), reporting in this paper the performance of several state-of-the-art LLMs using different prompt templates. Using LLMs as \oracles resulted in strong performance in the OAEI 2025, achieving the top-2 overall rank in the \emph{bio-ml} track.


\paragraph*{Supplemental Material:} 
 Source code and relevant resources for the experiments conducted in this paper are available in Zenodo: \url{https://doi.org/10.5281/zenodo.15394653}.
The source code for the experiments with different LLMs and prompts as diagnostic tools is available at \url{https://github.com/city-artificial-intelligence/rai-ukraine-kga-llm}.
The source code for LogMapLLM's integrated pipeline is available in this GitHub repository: \url{https://github.com/city-artificial-intelligence/logmap-llm}.

\end{abstract}

\section{Introduction}
\label{sec:intro}

Ontology alignment \cite{ombook2013} plays a crucial role in integrating diverse data sources across domains. While numerous ontology matching systems exist (\eg \cite{oaei2024}), systems capable of producing high-quality correspondences among the input ontologies are still needed, especially in applications where high confidence is paramount. 
One way to address this issue is through user interaction to manually verify uncertain mappings; however, this approach is often time-consuming and expensive. An alternative is to leverage Large Language Models (LLMs) as encoders of large amounts of data. LLMs have shown potential to be useful within an ontology alignment pipeline (\eg \cite{agentOM2024}). 
Nevertheless, LLMs are computationally or financially costly, and an unlimited use is not feasible.

In this paper, we have extended the state-of-the-art ontology matching system \logmap \cite{logmap2012} to perform calls to an LLM-based \oracle.
The LLM-based \oracle is used to validate a subset of correspondences where \logmap is uncertain. Thus, the LLM is invoked only for complex cases where traditional alignment techniques may be insufficient.
The calls to the LLM-based oracle are performed via ontology-driven prompts that exploit different levels of lexical and contextual information about the entities in the mappings in question. 
%
We selected the GPT-4o Mini model
(OpenAI) and a range of Google Gemini Flash models
for our experiments, due to their good performance in recent LLM leaderboards. 
%
%

To analyse the suitability of the LLM-based \oracles, we have conducted an extensive evaluation with the \emph{anatomy} \cite{anatomy2017}, \emph{largebio} \cite{largebio2012}, and \emph{bio-ml} \cite{bio-ml2022} datasets of the 
Ontology Alignment Evaluation Initiative (OAEI)~\cite{oaei2021,OAEI2025Results}, involving a total of nine matching tasks. These datasets are complex and have become a reference in the research community. 
We have assessed the diagnostic capabilities of thirty different LLM-prompt combinations based on the choice of five LLM implementations and six prompt templates. We have also evaluated the contribution of the LLM-based \oracles to the overall matching task by comparing the results with \logmap (automatic mode) and simulated \oracles with variable error rates \cite{uservalidationker2019}.
%
We also report experiments for the \textit{anatomy} dataset with the open-weight models Mistral, Llama and Qwen. 


In contrast with other state-of-the-art systems that rely heavily on LLMs, our approach is designed to only use the LLM-based \oracle in very specific cases. 
Hence, the use of LLMs is more accessible without the need for substantial computational infrastructure or financial resources.
The following points highlight the main contributions and novel aspects of this work. \textit{(i)}
 We investigate the effect of incorporating the ontology context of the entities into prompt design, an aspect that has not been thoroughly examined in the ontology alignment literature. \textit{(ii)}
 To our knowledge, while LLMs are increasingly applied in ontology alignment pipelines, their use as \Oracles has been unexplored in the state-of-the-art.
  \textit{(iii)} We provide a comprehensive evaluation that offers novel insights into the use of LLMs as diagnostic engines for ontology alignment, including a transparent and fine-grained analysis of the LLM contribution.
  \textit{(iv)}
  The combination of LogMap with an LLM-based \oracle achieved top-2 overall results in the OAEI 2025 \emph{bio-ml} track.

The paper is organised as follows. Section \ref{sec:prelim} introduces the necessary background. The relevant related work is provided in Section \ref{sec:related}. Section \ref{sec:methods} presents our method and system pipeline. Evaluation results are analysed in Section \ref{sec:eval}. Conclusions, future work and limitations are discussed in Section \ref{sec:conclusions} and Section \ref{sec:limitations}.
\section{Preliminaries}
\label{sec:prelim}




An ontology alignment is the process of finding correspondences or a \emph{mapping} $\M$ among the entities (ontology classes, properties or instances) of two or more ontologies. 
A \emph{mapping} involving two entities is typically represented as a 4-tuple $\mapping{e_1}{e_2}{r}{c}$ where $e_1$ and $e_2$ are entities of the ontologies $\On_1$ and $\On_2$, 
respectively, $r$ is a semantic relation, typically one of $\myset{\sqsubseteq, \sqsupseteq, \equiv}$, and $c$ is a confidence value (usually a number between $0$ and $1$).
For simplicity, in this paper, we refer to an equivalence mapping ($\equiv$) as a pair $\mappingTwo{e_1}{e_2}$.

\paragraph{Alignment task.} In the OAEI, an alignment or matching task is composed of a pair of ontologies, $\On_1$ (source) and $\On_2$ (target), and an associated \emph{reference alignment} $\MRA$. An $\MRA$, although it may not be perfect, serves as a guide to evaluating and comparing alignment systems.



\paragraph{Alignment system.}
An ontology \emph{alignment system} is a program
that, given as input an alignment task, 
generates an ontology alignment $\MS$.  We have selected the state-of-the-art alignment system \logmap \cite{logmap2012} as the baseline for our experiments due to its flexibility to be adapted to different evaluation scenarios. \logmap can operate in a fully automatic mode or allow interaction with an \emph{\Oracle} \cite{uservalidationker2019}. During the mapping selection stage, \logmap identifies a subset of mappings $\M_{ask}$ for which it is uncertain and would prefer to leverage the expertise of the \Oracle. 
If the \Oracle is not available, \logmap performs automatic decisions over $\M_{ask}$. 
\change{Appendix \ref{app:logmap} provides additional information about \logmap, including the workflow it follows when allowing interaction.}

\paragraph{Oracle.} We define an \Oracle as an external party that can assess the correctness of a given mapping $\mappingTwo{e_1}{e_2}$. An \Oracle can be a domain expert or an automated engine that exploits background knowledge. Additionally, the OAEI's interactive matching task simulates domain experts with different error rates via \Oracles relying on the reference alignment of the alignment task and randomly generating erroneous replies according to the selected error rate \cite{uservalidationker2019}.

\paragraph{Evaluation metrics.} 
We use the standard evaluation metrics \textit{Precision} (Pr), \textit{Recall} (Re), and \textit{F-score} (F) to evaluate an alignment $\MS$ computed by a system w.r.t. a reference alignment $\MRA$:

\begin{small}
\begin{equation*}\label{eq:measures}
    Pr = \frac{\lvert\MS \cap \MRA\rvert}{\lvert\MS\rvert},~
    Re = \frac{\lvert\MS \cap \MRA\rvert}{\lvert\MRA\rvert},~
    F = 2 \cdot \frac{Pr \cdot Re}{Pr + Re} 
\end{equation*}
\end{small}

We use \textit{Sensitivity} (Se), \textit{Specificity} (Sp), and \textit{Youden’s index} (YI) \cite{youdenindex1950}, as follows, to evaluate the effectiveness of an Oracle at diagnosing mappings in $\M_{ask}$, where TP, FN, TN, and FP stand for the usual true positive, false negative, true negative, and false positive counts, respectively, such that: 

\begin{small}
\begin{equation*}\label{eq:measuresS}
    Se = \frac{TP}{TP + FN},~
    Sp = \frac{TN}{FP + TN},~
    YI = Se + Sp - 1
\end{equation*}
\end{small}

%

\paragraph{LLM prompting.} LLMs like GPT-4 are pretrained on vast text corpora. They are commonly used in a few-shot or zero-shot setting via prompts. Prompts can exploit the generative capabilities of the LLM or ask for specific yes/no or True/False decisions. In the ontology alignment setting, 
a mapping $\mappingTwo{e_1}{e_2}$ can be transformed into a binary question to the LLM -- ``Does $e_1$ represent the same entity as $e_2$? (True/False)'' -- possibly enriched with ontology context (\eg parent classes or synonyms). This approach allows the LLM to be used as a lightweight semantic \oracle.




    



\section{Related Work}
\label{sec:related}
The Ontology Alignment Evaluation Initiative (OAEI)
has driven progress since 2004 by providing standardised benchmarks and evaluation protocols for matching systems \cite{oaei2024}. Widely-used traditional matchers include \logmap \cite{logmap2012} and AgreementMakerLight (AML) \cite{aml-2025}, each leveraging different combinations of lexical, structural, and background‐knowledge techniques.
Human validation has long been recognised as critical for high-precision mappings. Early 
frameworks combined automated matching with domain expert feedback to resolve low-confidence correspondences, but at the cost of extensive user effort and time \cite{uservalidationker2019}. 

In recent years, a new generation of systems leveraging Machine Learning (ML) and (large) language models has emerged. The OAEI Bio-ML track \cite{bio-ml2022} was established to foster participation in the OAEI and to facilitate the systematic evaluation of these systems.
Early approaches showed promising results applying word embeddings to the ontology alignment task (\eg \cite{kolyvakis2018deepalignment,nkisi2018ontology,veealign2021}). Knowledge graph embedding systems like OWL2Vec*~\cite{owl2vec2021} were also deployed in combination with ML to learn and validate ontology alignment (\eg \cite{logmapml2021,Onto-ML-JWS2023}). Systems relying on BERT-based models have become popular, given their flexibility to fine-tuning for specific tasks like ontology alignment. Prominent examples include BERTMap \cite{bertmap2022}, BioGITOM~\cite{BioGITOM2024}, and the Matcha family \cite{matcha2023}. Recent developments in the field are increasingly driven by approaches based on LLMs. Saki Norouzi et al. \cite{convOA2023} and He et al. \cite{exploringllm4oa2023} performed exploratory studies about the potential of LLMs at ontology alignment. Amini et al. \cite{complexOALLM2024} extended the exploration to discover complex alignments beyond equivalence or subsumption. 
Systems like OLaLa~\cite{olala2023}, LLMs4OM \cite{llms4oa2024}, MILA \cite{MilaTaboada2025}, Agent-OM \cite{agentOM2024}, KROMA \cite{kroma2025} and HybridOM \cite{HybridOM2024} have integrated LLMs in their architectures. A common technique has been to use retrieval methods to select top-k candidates for each entity, then asking the LLM to select the best among these candidates (\eg \cite{olala2023,llms4oa2024,MilaTaboada2025}). By contrast, HybridOM uses an LLM to generate additional lexical descriptions of the entities involved in candidate correspondences.
Recent approaches have explored the use of LLMs to focus on the alignment of 
instance data within knowledge graphs (\eg \cite{autoalign2023,glam2024}).
Agent-OM proposed the use of autonomous LLMs to orchestrate multiple matching subtasks, indicating potential for agentic AI workflows to be adopted for ontology matching \cite{agentOM2024}.

Our approach builds upon \logmap \cite{logmap2012} and employs the LLM as an \oracle to assess a targeted subset of mappings. Rather than attempting to evaluate a large set of candidate correspondences, we focus on the validation of mappings where \logmap is uncertain.
\change{Systems like MILA \cite{MilaTaboada2025} and KROMA \cite{kroma2025} have also focused on the limitation of the number of queries, leading to a reduction in the required computation times.} 


\section{Methods: LLMs as \oracle}
\label{sec:methods}

\begin{figure}[t!]
\vspace{-0.1cm}
    \centering
    \includegraphics[width=0.7\linewidth]{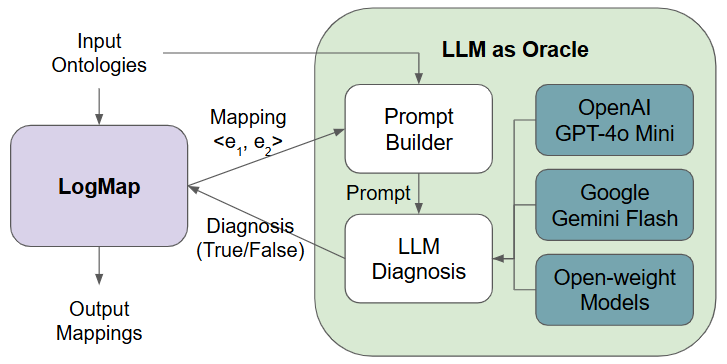}
    \caption{\change{LLM-in-the-loop as an 
    \Oracle to diagnose challenging matches in ontology alignment.}}
    \label{fig:llm-oracle}
\end{figure}

As introduced in Section \ref{sec:prelim}, we build upon the system \logmap. In addition to predicting a set of output mappings, \logmap also identifies a subset of uncertain mappings ($\Mask$), which 
can optionally be given to an \Oracle.
In this paper, we have extended the architecture of \logmap to use a state-of-the-art \emph{LLM as an \Oracle} as depicted in Figure \ref{fig:llm-oracle}. 
We restrict the use of the LLM to the mappings in $\Mask$. 
These mappings are not trivial as they typically involve entities with different labels and/or contexts, and they are better suited as a challenge to the performance of LLMs.
\logmap interacts with the \Oracle on-demand for each mapping $\mappingTwo{e_1}{e_2} \in \Mask$. 
The following subsections detail the internal steps involved in the interaction with the LLM-based \Oracle.


\subsection{Ontology-driven prompt builder}
\label{sec:templates}

The first step in the interaction with the LLM-based \Oracle is the creation of an ontology-driven prompt to ask about the correctness of a given candidate mapping $\mappingTwo{e_1}{e_2}$. 
%
The ontologies provide lexical representations (\eg labels and synonyms), as well as context (\eg parent classes) for $e_1$ and $e_2$.
According to the locality principle \cite{umlsassessmen2011}, mappings should link
entities that have similar contexts. 
Hence, a basic prompt should include at least the lexical representation of entities $e_1$ and $e_2$, and that of one of their directly connected entities.

We have designed six different prompt templates combining three characteristics: \textit{(i)}~using similar sentences to how humans write (natural language-friendly, \textbf{NLF}), \textit{(ii)}~inclusion of extended context (\textbf{EC}), and \textit{(iii)} inclusion of synonyms (\textbf{S}). 
Prompts without an extended context only include one of the direct parents for classes and properties, and one of the direct types for individuals. While the prompts with an extended context include two levels of parent classes. We also evaluate combinations of the above characteristics. 
We refer to as \PECSNLF the prompt using all \textbf{NLF}, \textbf{EC} and \textbf{S} characteristics. 
 
%
%
For each mapping $\mappingTwo{e_1}{e_2}$ to be assessed, we dynamically populate each of the prompt templates according to the entities in the mapping and their associated ontology information.  Below, we show the populated prompts for the mapping 
$\langle$\texttt{mouse:MA\_0001771} (alveolus epithelium), \texttt{human:NCI\_C12867} (Alveolar\_Epithelium) $\rangle$.

\paragraph{Structured prompts.} 
This type of prompt uses structured information with uncommon natural language expressions. Listing~\ref{lst:basicprompt} shows the \Prompt prompt, where the entities and their context are listed. \change{Listing~\ref{lst:pec} in Appendix~\ref{app:prompts} shows the structured prompt with extended context~\PEC.}


\begin{lstlisting}[basicstyle=\scriptsize\ttfamily, caption={Basic prompt without any of the characteristics enabled (\Prompt) },%
                   label={lst:basicprompt}]
Analyze the following entities, each originating from a distinct ontology. Your task is to assess whether they represent the **same ontological concept**, considering both their semantic meaning and hierarchical position.

1. Source entity: "alveolus epithelium"
    - Direct ontological parent: lung epithelium

2. Target entity: "Alveolar_Epithelium"
    - Direct ontological parent: Epithelium

Are these entities **ontologically equivalent** within their respective ontologies? Respond with "True" or "False".
\end{lstlisting}


\paragraph{Natural-language friendly prompts.} 
These prompts are based on the assumption that, given that LLMs are trained on large corpora of human-generated text, formulating questions in a more human-like way is expected to yield more accurate results. \change{Listing \ref{lst:prompt_nlf} shows this type of  
prompt (\PNLF), while Listing \ref{lst:prompt_ec_nlf} in Appendix \ref{app:prompts} includes the version with extended context \PECNLF.}


\begin{lstlisting}[basicstyle=\scriptsize\ttfamily,caption={\PNLF Prompt (natural-language friendly).},%
                   label={lst:prompt_nlf}]
We have two entities from different ontologies.

The first one is "alveolus epithelium", which belongs to the broader category "lung epithelium"

The second one is "Alveolar_Epithelium", which belongs to the broader category "Epithelium"

Do they mean the same thing? Respond with "True" or "False".
\end{lstlisting}

\paragraph{Prompts with synonyms.} Although LLMs may inherently encode synonyms and lexical variations related to the ontology entities, the \PSNLF and \PECSNLF prompts are designed to analyse the impact of explicitly including synonyms for both the entities in a given correspondence and their associated context. \change{A \PSNLF prompt is shown in Listing~\ref{lst:prompt_nlf_s}, while Listing~\ref{lst:prompt_nlf_s_ec} in Appendix~\ref{app:prompts} provides its variant with extended context (\PECSNLF).}



\begin{lstlisting}[basicstyle=\scriptsize\ttfamily,caption={\PSNLF Prompt (natural-language friendly with synonyms).},%
                   label={lst:prompt_nlf_s}]
We have two entities from different ontologies.

The first one is "alveolus epithelium", which falls under the category "lung epithelium".

The second one is "Alveolar_Epithelium", also known as "Lung Alveolar Epithelia", "Alveolar Epithelium", "Epithelia of lung alveoli", which falls under the category "Epithelium".

Do they mean the same thing? Respond with "True" or "False".
\end{lstlisting}



\paragraph{System prompts.} In addition to the above mapping templates, it is possible to add, for each LLM session, a short message that frames the model's overall role and answering style before it sees any individual mapping question. We experimented with sessions using no system prompt as well as various system prompt variants, positioning the LLM as follows:
\textit{(i)} as an ontology matching expert to ensure precision (\textit{base}); 
\textit{(ii)} to explain its decision in a natural-language friendly manner (\textit{explainable}); 
\textit{(iii)} emphasizing the use of hierarchical and semantic context (\textit{hierarchical}); and 
\textit{(iv)} to leverage explicitly provided synonyms and parent‐class semantics (\textit{lexical}). 
\change{Listing \ref{lst:systemprompt} in Appendix \ref{app:prompts} includes the specific system prompts.}










\setlength{\tabcolsep}{2.35pt}
\begin{table}[t!]
    \centering
    \resizebox{\columnwidth}{!}{
    \begin{tabular}{|l|c|c|r|r|r|r|}
        \hline
\multirow{2}{*}{\textbf{Model}} & \multicolumn{2}{c|}{\textbf{Cost / 1M Tokens}} & \multicolumn{2}{c|}{\textbf{Request Limits}}
& \multicolumn{1}{c|}{\textbf{Cost}} & \multicolumn{1}{c|}{\multirow{2}{*}{ \textbf{Latency (s)}}} \\\cline{2-5}
 & \textbf{~~Input~~} & \textbf{Output} & \multicolumn{1}{c|}{\textbf{~Minute~}} & \multicolumn{1}{c|}{\textbf{Day}} & \multicolumn{1}{c|}{\textbf{1k requests}} &  \\ \hline\hline

\textbf{Qwen3-8b (local)}  & - & - & 0.5 & $<$1,000 & - & $>$125 \\\hline

\textbf{Mistral Small-2402} & \$1.00 & \$3.00 & 400 & 576,000 & \$0.15--\$0.23  & 6--10 \\\hline
\textbf{Llama 3-70b} & \$2.65 & \$3.50 & 800 & 1,152,000 & \$0.35--\$0.62 & 7--20 \\\hline\hline

\textbf{Gemini 1.5 Flash}& \$0.08 & \$0.30 & 2,000 & 2,880,000 & \$0.010--\$0.018  & 6--10  \\\hline

\textbf{Gemini 2.0 Flash}& \$0.10 & \$0.40 & 2,000 & 2,880,000 & \$0.014--\$0.024  & 4--7  \\\hline

\textbf{Gemini 2.0 Flash-Lite}& \$0.08 & \$0.30 & 4,000 & 5,760,000 & \$0.010--\$0.018 & 5.5--7.5  \\\hline

\textbf{Gemini 2.5 Flash}& \$0.15 & \$0.60 & 1,000 & 10,000 & \$0.018--\$0.033 & 6.5--8  \\\hline

\textbf{GPT-4o Mini}& \$0.15 & \$0.60 & 500 & 10,000 & \$0.025--\$0.04 & 4--14  \\\hline
\end{tabular}
}
\caption{Latency and cost of evaluated LLMs. Each request typically consumed between 100 and 250 input tokens and 5 to 10 output tokens.}
\label{tab:latency_cost}
\end{table}

\subsection{LLM-based diagnosis}
\label{sec:llm}

Our selected LLMs include GPT-4o Mini (OpenAI) and a range of Google Gemini Flash models (v1.5, 2.0, 2.0 Lite, and 2.5 Preview). These models were chosen based on their balance of cost-effectiveness, response latency, scalability, reliability, and output quality, as compared to other commercial APIs and open-weight alternatives. Furthermore, these LLMs expose a consistent client interface, enabling straightforward integration into our system.
Support for lightweight models such as GPT-4o Mini and Gemini 2.0 Flash-Lite ensures accessibility for researchers operating under constrained budgets. At the same time, including a progression of Gemini Flash versions (from v1.5 to v2.5) allows us to observe how model improvements over time impact diagnostic performance in the ontology alignment task.


In order to achieve binary (True/False) diagnostic classification, we 
used a structured output feature. We define a Boolean answer that will be a decider of the zero-shot question that we ask the LLM. \change{To enhance robustness, we incorporated a validation and retry mechanism, that is, if the output is not parsed correctly (e.g., neither True nor False), we resend the same request.}
%

\change{
We used the Chat Completions API~\cite{chat2025} for 
GPT-4o Mini,
and the OpenAI's SDK endpoint for the Gemini Models \cite{geminiAPI}. Response latency typically 
remains within a few seconds, 
depending on prompt complexity and model characteristics. 
This enabled high-throughput querying, especially when requests were executed in parallel.
However, API rate limits imposed practical constraints on experimentation. The Gemini API permits up to 2,000 requests per minute (RPM) by default \cite{geminiLimits}, whereas OpenAI's API begins with a limit of 500 RPM and a daily quota of 10,000 requests \cite{openAILimits}, thereby restricting the overall throughput of our experiments.
Regarding the cost of experiments, token usage is a key factor. Each request typically consumes between 100 and 250 input tokens, depending on the complexity and detail of the prompt. Pricing per million input tokens varied per model \cite{openAILimits,geminiPricing}.
The average cost per 1,000 requests ranged from approximately \$0.01~to~\$0.04. Table~\ref{tab:latency_cost} summarises the cost and latency of the evaluated LLM models.}


\paragraph{Open-weight models.} 
\change{Our end-to-end evaluation focuses on commercial LLMs due to their ease of integration via their APIs and cost-effective performance. This choice, however, may limit reproducibility and accessibility for users or institutions that prefer or require open-weight alternatives. Hence, we also performed a preliminary evaluation with the open-weight models Mistral, Llama (Meta), and Qwen (Alibaba Cloud).
Mistral Small (2402, approx. 24b) and Llama 3-70b Instruct were accessed via the Amazon Bedrock API \cite{bedrockmistral,bedrockllama}. Mistral Small typically responded to mapping requests in less than 10 seconds, while Llama 3-70b took between 7 and 20 seconds. 
The cost per 1,000 requests was under \$0.23 for Mistral Small-2402 and \$0.62 for Llama3-70b.  
Qwen3 models (1.7b and 8b) \cite{qwen3} were run locally on a standard laptop equipped with an integrated M2 GPU. Due to the limitations of the local setup, the average latency per request exceeded 125 seconds. Table \ref{tab:latency_cost} also summarises the cost and latencies for the evaluated open-weight models.
}

\subsection{Impact of the Oracle}
\label{sec:oracleimpact}

The diagnosis performed by the \Oracle over the mapping set $\Mask$ may have an impact on the overall \logmap performance as it may lead to the acceptance or rejection of additional mappings. The authors in \cite{uservalidationker2019} simulated \Oracles with different error rates and performed an extensive analysis of the impact and error propagation of the Oracle decisions. In this work, we have followed a similar approach to evaluate the LLM-based \Oracle (\OrLLM) against \Oracles with error rates ranging from 0\% (\ie perfect \oracle, \OrZero) to 30\% (\ie \OrThirty).
The simulated \Oracles rely on the reference alignment of the relevant matching task and generate erroneous replies with the probability of their associated error rate. These \Oracles with uniformly distributed errors do not realistically represent how a domain expert would behave, but they serve our purpose to assess the performance of the LLM-based \Oracle in comparison with potential domain experts that are likely to make mistakes \cite{uservalidationker2019}.

\section{Experimental evaluation}
\label{sec:eval}

\setlength{\tabcolsep}{2pt}
\begin{table}[tb]
    \centering
    \begin{tabular}{|l|l||c|c|c|}
        \hline
        \textbf{OAEI track} &
        \textbf{Matching task} &  $\lvert\ontoOne\rvert$ & $\lvert\ontoTwo\rvert$ & $\lvert\MRA\rvert$\\ \hline\hline 
        
         \multirow{1}{*}{\textbf{Anatomy}} 
         & \textbf{Mouse-Human} & 2,755 &3,313 & 1,516   \\ \hline\hline
         \multirow{5}{*}{\textbf{Bio-ML}}         
         & \textbf{NCIT-DOID} & 15,991 & 8,516 & 4,686   \\ \cline{2-5}
         & \textbf{OMIM-ORDO} & 9,662 & 9,320 &3,721   \\ \cline{2-5}
         & \textbf{SNOMED-FMA.body} & 34,562 & 89,180 & 7,256   \\ \cline{2-5}
         & \textbf{SNOMED-NCIT.neoplas} & 23,116 & 20,497 & 3,804   \\ \cline{2-5}
         & \textbf{SNOMED-NCIT.pharm} & 29,646 & 22,387 & 5,803
   \\ \hline\hline
         \multirow{3}{*}{\textbf{Largebio}}
         & \textbf{FMA-NCI} & 79,049 & 66,919 & 3,024
   \\ \cline{2-5}
         & \textbf{FMA-SNOMED} & 79,049 & 122,521 & 9,008
   \\ \cline{2-5}
         & \textbf{SNOMED-NCI} & 122,521 & 66,919 & 18,844  \\ \hline
     \end{tabular}
     \caption{Statistics of the used OAEI datasets. Ontology size is given in terms of the number of entities. $\MRA$ is the reference alignment of the matching task.}    
     \label{tab:stats} 
\end{table}

Our experiments were conducted on a standard laptop with the selected LLM models as detailed in Section \ref{sec:llm}.
All the experiments reported here were obtained with a budget of less than \$50.
We used the \emph{anatomy} \cite{anatomy2017}, \emph{largebio} \cite{largebio2012}, and \emph{bio-ml} \cite{bio-ml2022} datasets provided by the OAEI evaluation initiative \cite{oaei2021,oaei2024}. 
As shown in Table~\ref{tab:stats}, we covered a total of nine ontology matching tasks, involving ontologies of diverse sizes containing mostly concepts. 
The reference alignments ($\MRA$) of these matching tasks have different sources. In \emph{anatomy}, the reference alignment has been manually curated, while in \emph{bio-ml} and \emph{largebio} the reference alignment relies on 
public resources like MONDO \cite{mondo} and UMLS~\cite{umls}.

\begin{figure*}[t!]
    \centering
    \includegraphics[width=0.999\linewidth]{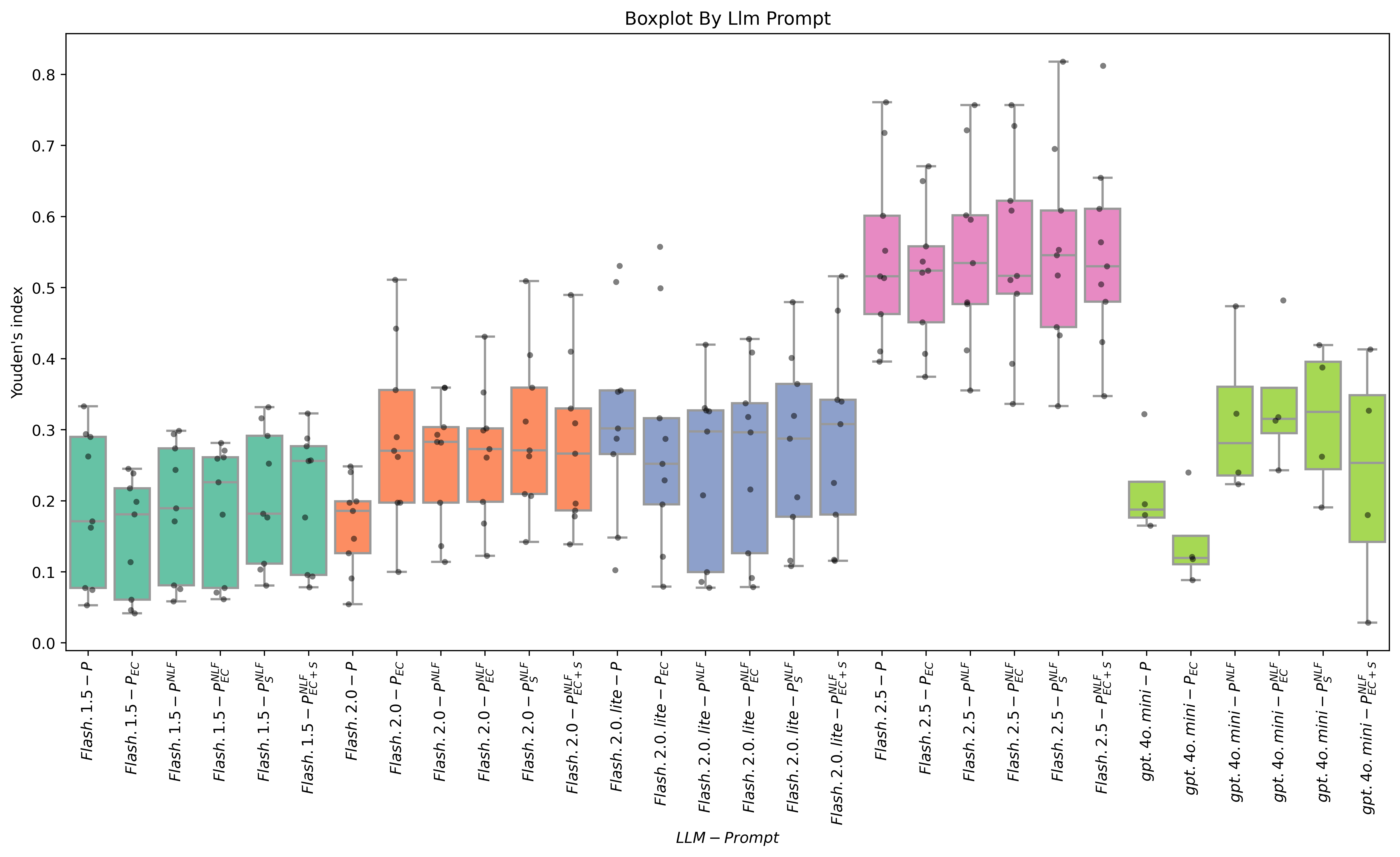}
    \caption{Summary of the diagnostic results (Youden’s index) for the LLM-based \Oracles. 
    }
    \label{fig:all-results-plot}
\end{figure*}

\paragraph{Diagnostic capability.} We tested over the 9 matching tasks a total of 30 LLM-based \Oracles (\OrLLM), combining the six prompt templates introduced in Section \ref{sec:templates} with the LLM models referred to in Section \ref{sec:llm}.
Figure \ref{fig:all-results-plot} shows the Youden’s index (YI) as a measure of the correctness of the LLM-based \oracles for each LLM and prompt template combination. \change{Detailed results per matching task are provided in Appendix \ref{app:results}.}

\Oracles relying on the Gemini 2.5 Flash model led to the best results on average, as summarised in Figure~\ref{fig:all-results-plot}. The best results were achieved by the combination of Gemini~2.5 Flash and \PSNLF prompts, which we refer to as \OrGemNew. 
Table \ref{tab:correctnessoracle} compares the performance of \logmap (automatic mode) and the best model combination \OrGemNew in diagnosing the mappings in $\Mask$. As anticipated, \logmap performs poorly as a diagnostic engine for the mappings in $\Mask$, yielding YI values close to 0 (\ie no discriminative power), whereas \OrGemNew achieves significantly better results, with an average YI value exceeding 0.5. 

The YI index captures the effectiveness of an \Oracle at identifying positive (sensitivity) and negative (specificity) mappings.
A YI value of 1.0 indicates optimal performance. %
\change{While no standard cut-off values exist
for YI, some papers use 0.3, 0.5 and 0.7 as representative values for low, moderate and high effectiveness, respectively (\eg \cite{youdenindexcutoffs}).
Due to the complexity of the mappings in $\Mask$, moderate YI values can be expected of an \oracle.}

\setlength{\tabcolsep}{3.25pt}
\begin{table}[t!]
    \centering
    \resizebox{\columnwidth}{!}{
    \begin{tabular}{|l||c|c||c|c|c||c|c|c|}
        \hline
        \multirow{2}{*}{\textbf{Matching task}} & \multicolumn{2}{c||}{$\lvert\Mask\rvert$}  & \multicolumn{3}{c||}{\textbf{\logmap on $\Mask$}} & \multicolumn{3}{c|}{\textbf{\OrGemNew on $\Mask$}}   \\ \cline{2-9}  
        &
        \textbf{P} & \textbf{N} & \textbf{~~~Se~~~} & \textbf{~~~Sp~~~} & \textbf{~~~YI~~~} & \textbf{~~~Se~~~} & \textbf{~~~Sp~~~} & \textbf{~~~YI~~~} \\ \hline\hline
        \textbf{Mouse-Human} & 165 & 94 & 1.000 & 0.000 & 0.000 & 0.951 & 0.744 & \textbf{0.695}  \\ \hline\hline
        \textbf{NCIT-DOID} & 364 & 492 & 1.000 & 0.000 & 0.000 & 0.849 & 0.697 & \textbf{0.546} \\ \hline
        \textbf{OMIM-ORDO} & 172 & 227 & 0.343 & 0.551 & -0.106 & 0.942 & 0.876 & \textbf{0.818} \\ \hline        
        \textbf{SNOMED-FMA.body} & 369 & 619 & 0.881 & 0.149 & 0.029 & 0.884 & 0.669 & \textbf{0.553} \\\hline
        \textbf{SNOMED-NCIT.neoplas} & 704 & 601 & 0.984 & 0.067 & 0.051 & 0.840 & 0.593 & \textbf{0.433} \\\hline
        \textbf{SNOMED-NCIT.pharm} & 297 & 260 & 0.929 & 0.065 & -0.005 & 0.848 & 0.669 & \textbf{0.517} \\\hline\hline
        \textbf{FMA-NCI} & 410 & 475 & 0.705 & 0.726 & 0.431 & 0.761 & 0.684 & \textbf{0.445} \\\hline
        \textbf{FMA-SNOMED} & 831 & 621 & 0.941 & 0.225 & 0.166 & 0.480 & 0.853 & \textbf{0.333} \\\hline
        \textbf{SNOMED-NCI} & 1450 & 1128 & 0.887 & 0.395 & 0.281 & 0.846 & 0.763 & \textbf{0.609} \\\hline\hline

        \textbf{Average} & 529 & 502 & 0.852 & 0.242 & 0.094 & 0.822 & 0.728 & \textbf{0.550} \\\hline

     \end{tabular}
    }
    \caption{Comparison of \logmap (automatic mode) against the best LLM-based \Oracle (\OrGemNew, using Gemini 2.5 Flash and \PSNLF prompts) to diagnose the correctness of $\Mask$. P is the number of real positives in $\Mask$, N the number of real negatives, Se denotes Sensitivity, Sp Specificity and YI the Youden’s index.}    
    \label{tab:correctnessoracle} 
\end{table}

\paragraph{Impact of the prompt template.} 
Figure \ref{fig:all-results-plot} also illustrates the impact of using different prompt templates. 
Results vary across LLM models. Natural-language friendly prompts produce more consistent behaviour, while incorporating extended context and synonyms has a positive impact.
Overall, for the Gemini 2.0 Flash and 2.5 Flash models, the most effective prompts were \PSNLF (natural-language friendly with synonyms).  
\change{We emphasise that different ontologies and matching tasks pose varying challenges due to lexical and structural differences (see Appendix \ref{app:results} for an overview of the results per matching task). Our tested prompts aim to capture this by leveraging both structural and lexical information in the input ontologies. 
There is also a dependency on the selected $\Mask$ mappings by \logmap, which may also be more complex in some tasks than others (\eg mappings involving isolated entities and/or with scarce synonyms).}

\setlength{\tabcolsep}{0.5pt}
\begin{table}[t!]
    \centering
    \resizebox{\columnwidth}{!}{
    \begin{tabular}{|l||c|c|c||c|c|c||c|c|c|}
        \hline
        \multirow{2}{*}{\textbf{Matching task}} & \multicolumn{3}{c||}{\textbf{\logmap}} & 
        \multicolumn{3}{c|}{\textbf{\logmap ~- \OrGem}} &
        \multicolumn{3}{c|}{\textbf{\logmap ~- \OrGemNew}}   \\ \cline{2-10}  
        &
        \textbf{Pr} & \textbf{Re} & \textbf{F} & 
        \textbf{Pr} & \textbf{Re} & \textbf{F} & 
        \textbf{Pr} & \textbf{Re} & \textbf{F}  \\\hline\hline
        \textbf{Mouse-Human} & ~0.915~ & ~0.848~ & ~0.880~ & ~0.945~ & ~0.844~ & ~0.892~ & ~0.963~ & ~0.842~ & \textbf{~0.898~}  \\ \hline\hline
        \textbf{NCIT-DOID} & 0.845 & 0.895 & 0.869 & 0.875 & 0.890 & 0.882 & 0.907 & 0.883 & \textbf{0.895}  \\ \hline
        \textbf{OMIM-ORDO} & 0.874 & 0.448 & 0.592 & 0.882 & 0.478 & 0.620 & 0.914 & 0.476 & \textbf{0.626}
 \\ \hline
        \textbf{SNOMED-FMA.body} & 0.695 & 0.538 & 0.607 & 0.727 & 0.543 & 0.622 & 0.751 & 0.545 & \textbf{0.632} \\\hline
        \textbf{SNOMED-NCIT.neoplas} & 0.624 & 0.774 & 0.691 & 0.636 & 0.763 & 0.694 & 0.661 & 0.747 & \textbf{0.701} \\\hline
        \textbf{SNOMED-NCIT.pharm} & 0.825 & 0.625 & 0.711 & 0.847 & 0.625 & \textbf{0.719} & 0.855 & 0.621 & \textbf{0.719} \\\hline\hline
        \textbf{FMA-NCI} & 0.860 & 0.800 & 0.829 & 0.901 & 0.796 & \textbf{0.845} & 0.853 & 0.804 & 0.828 \\\hline
        \textbf{FMA-SNOMED} & 0.796 & 0.641 & 0.710 & 0.814 & 0.644 & \textbf{0.719} & 0.854 & 0.585 & 0.694 \\\hline
        \textbf{SNOMED-NCI} & 0.868 & 0.650 & 0.743 & 0.866 & 0.656 & 0.747 & 0.897 & 0.646 & \textbf{0.751} \\\hline\hline
        \textbf{Average} & 0.811 & 0.691 & 0.737 & 0.833 & 0.693 & \textbf{0.749} & 0.851 & 0.683 & \textbf{0.749} \\\hline
        
     \end{tabular}
    }
    \caption{Comparison of \logmap (automatic mode) with \logmap with the best LLM-based \Oracles (\OrGem and \OrGemNew) on all matching tasks. Pr denotes Precision, Re Recall and F is the F-score.} 
    \label{tab:fulltask} 
    \vspace{-0.4cm}
\end{table}


\begin{figure}[t!]
    \centering
    \includegraphics[width=1.0\linewidth]    {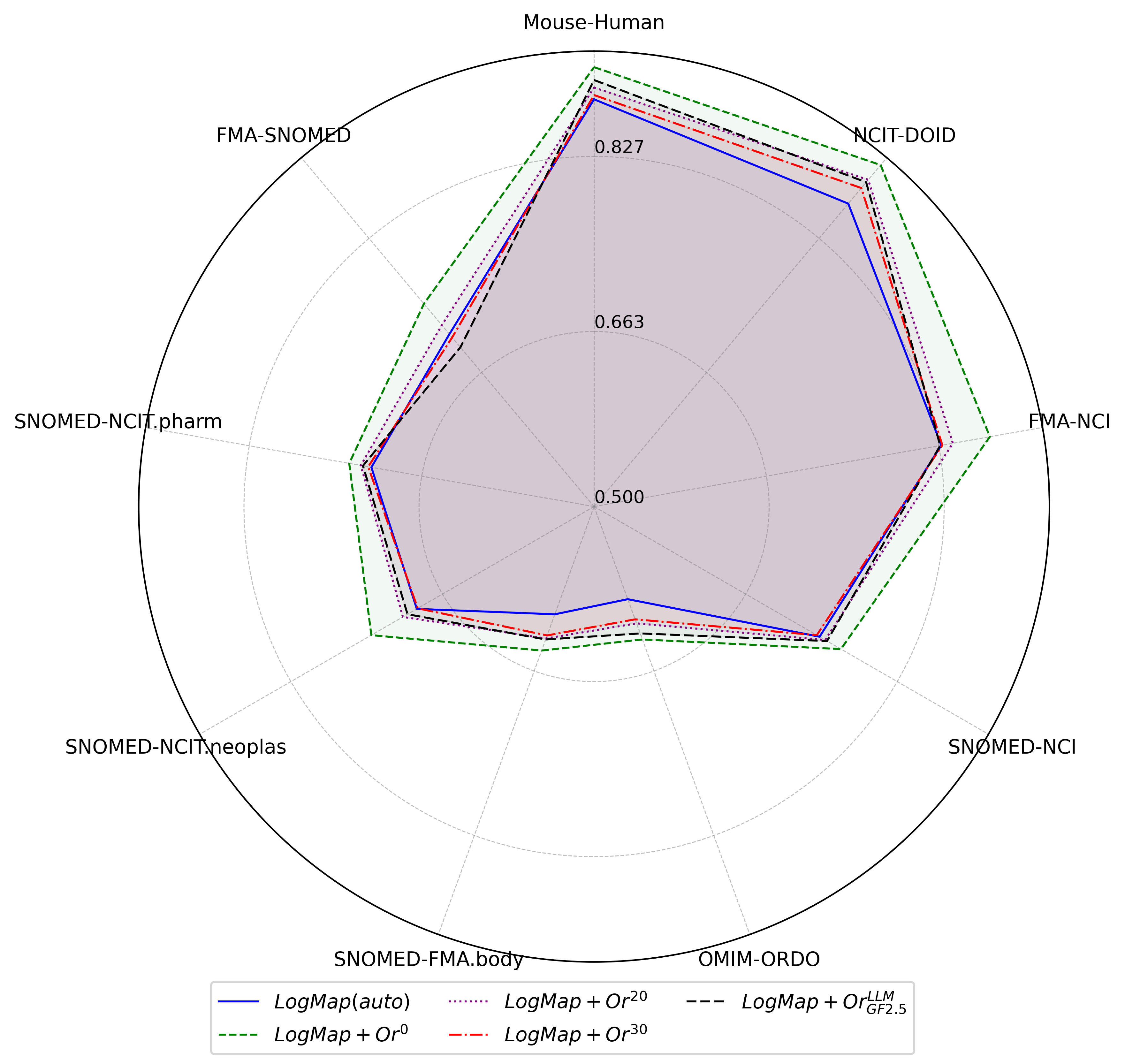}
    \caption{Comparison of \logmap, \logmap with \OrGemNew, and \logmap in combination with Oracles with different error rates (\OrZero, \OrTwenty, and \OrThirty).}
    \label{fig:radar-error-rate}
\end{figure}

\paragraph{Overall matching task.} 
Table \ref{tab:fulltask} shows the results obtained using \logmap (automatic mode)
compared to \logmap integrated with an LLM-based \Oracle (as depicted in Figure \ref{fig:llm-oracle}). 
We selected the top-performing LLM-based \Oracles using \PSNLF prompts: \OrGem (based on Gemini 2.0 Flash) and \OrGemNew (based on Gemini 2.5 Flash).
As anticipated, the integration with the LLM-based \Oracle yields improved F-scores across all tasks. \logmapP\OrGemNew dominates the \emph{anatomy} and \emph{bio-ml} tasks, while \logmapP\OrGem achieves the best results on \emph{largebio}. To better contextualise the effectiveness of the LLM-based \Oracle, we compared performance also with simulated \Oracles with various error rates, following the approach in \cite{uservalidationker2019} as detailed in Section \ref{sec:oracleimpact}. Figure \ref{fig:radar-error-rate} compares performance across all nine ontology matching tasks for:
\logmap, \logmapP\OrGemNew, and \logmap with the simulated Oracles \OrZero, \OrTwenty, and \OrThirty (corresponding to error rates of 0\%, 20\%, and 30\%, respectively). We can observe that \OrGemNew performs similarly to \OrTwenty, except in the FMA-SNOMED task (lower F-score) and OMIM-ORDO task (higher F-score). In line with previous studies \cite{logmap2012,uservalidationker2019}, \logmapP\OrThirty
still outperforms \logmap (without \Oracle). \change{The statistical analysis in Appendix \ref{app:stat} supports these observations.}

\paragraph{Comparison with OAEI systems.}
The results of \logmapP\OrGemNew are highly competitive when compared with the state-of-the-art systems participating in the OAEI campaign (see results in the OAEI 2021 \cite{oaei2021} for \emph{largebio}, and in the OAEI 2024 \cite{oaei2024}, and OAEI 2025 \cite{OAEI2025Results} for \emph{anatomy} and \emph{bio-ml}). For example, \logmapP\OrGemNew would have ranked top-3 in the 2024 \emph{anatomy} track, top-2 in the 2021 \emph{largebio} track, achieving performance comparable to leading systems such as BertMap \cite{bertmap2022}, Matcha \cite{matcha2023}, and LogMap-Bio \cite{logmapbioportal2014} in the 2024 \emph{bio-ml} track. 
In the OAEI~2025, \logmapP\OrGemNew participated under the name \logmapllm, achieving top-4 results in the \emph{anatomy} track and top-2 results in the \emph{bio-ml} track~\cite{OAEI2025Results,LogMapLLM2025}.
Table~\ref{tab:oaei-logmapllm} reports the official OAEI results and ranks achieved by \logmap and \logmapllm in the \emph{bio-ml} track.\footnote{
Note that the F-scores in Table~\ref{tab:fulltask} differ from the official OAEI \emph{bio-ml} results, as this track does not consider the complete ground truth for (global matching) evaluation. Further details are available at \url{https://liseda-lab.github.io/OAEI-Bio-ML/2025/index.html}.} 
LogMap-Bio\footnote{LogMap-Bio \cite{logmapbioportal2014} uses BioPortal as a source of mediating (biomedical) ontologies.} was, on average, the top performer in the \emph{bio-ml} track with an (average) F-score of 0.762, closely followed by \logmapllm with an (average) F-score of 0.751.

\setlength{\tabcolsep}{2.5pt}
\begin{table}[t!]
    \footnotesize
    \centering
    \resizebox{\columnwidth}{!}{
    \begin{tabular}{|l||c|c|c|c||c|c|c|c|}
        \hline
        \multirow{2}{*}{\textbf{Matching task}} & \multicolumn{4}{c||}{\textbf{\logmap}} & 
        \multicolumn{4}{c|}{\textbf{\logmapllm}}    \\ \cline{2-9}  
        &
        \textbf{Pr} & \textbf{Re} & \textbf{F}  & \textbf{Rank} &       
        \textbf{Pr} & \textbf{Re} & \textbf{F}  & \textbf{Rank} 
          \\\hline\hline
        \textbf{NCIT-DOID} & 
        0.843 & 0.893 & 0.867 & \#4 &
        ~0.932~ & ~0.883~ & ~0.907~ & \#2\\ \hline    
        \textbf{OMIM-ORDO} & 
        0.834 & 0.456 & 0.589 & \#7 &
        ~0.916~ & ~0.476~ & ~0.626~ & \#4 \\ \hline
       
        \textbf{SNOMED-FMA.body} &        
        0.760 & 0.569 & 0.651 & \#6 &
        ~0.869~ & ~0.561~ & ~0.682~ & \#4 \\ \hline

        \textbf{SNOMED-NCIT.neoplas} &         
        0.763 & 0.772 & 0.736 & \#4 &
        ~0.821~ & ~0.747~ & ~0.782~ & \#1 \\ \hline
        
        \textbf{SNOMED-NCIT.pharm} &
        0.932 & 0.620 & 0.745 & \#4 &
        ~0.979~ & ~0.621~ & ~0.760~ & \#1 \\ \hline\hline
        \textbf{OVERALL} & 0.826 & 0.662 & 0.718 & \#5 & 0.903 & 0.658 & 0.751 & \#2\\\hline
     \end{tabular}
    }
    \caption{\logmap and \logmapllm in the OAEI 2025 \emph{bio-ml} track. Pr=Precision, Re=Recall, F=F-score. Rank represents the position out of 10 participants.} 
    \label{tab:oaei-logmapllm} 
\end{table}

\paragraph{Determinism of the LLM-based Oracles.} 
The reliability of systems built on LLMs is a critical concern. Thus, we assessed the variability in the performance of the LLM-based Oracle across multiple independent runs, as well as the influence of the system prompt/message (detailed in Section \ref{sec:templates}). In this experiment, we used the Gemini 2.0 Flash and Flash-Lite models, applying all six prompt templates across three matching tasks. Performance variation over four separate runs was negligible \change{(\ie the observed standard deviation for YI ranged from $0.001$ to $0.005$, see Appendix~\ref{app:det} for details).}
\change{While system prompts did not lead to significant changes, they had a modest impact on 
performance, suggesting that framing the LLM context is important (see Appendix~\ref{app:det}).}

\paragraph{Opportunities with Open-weight models.}

\setlength{\tabcolsep}{1.25pt}
\begin{table}[t!]
    \centering
    \begin{tabular}{|l||c|c|c|}
        \hline
        \textbf{LLM Model} & \textbf{Sensitivity} & \textbf{Specificity} & \textbf{Youden's Index} \\ \hline\hline

       \textbf{Mistral Small-2402}  & 0.945 & 0.547 & 0.492\\\hline
       \textbf{Llama 3-70b} & 0.989 & 0.359 & 0.348 \\\hline
        \textbf{Qwen3-1.7b} & 0.811 & 0.411 & 0.222 \\\hline
        \textbf{Qwen3-8b} & 0.764 & 0.825 & \textbf{0.590}  
        \\\hline\hline
        
        \textbf{Gemini 1.5 Flash} & 0.994 & 0.322 &	0.316 \\\hline
        \textbf{Gemini 2.0 Flash} & 0.994 & 0.411 & 0.405 \\\hline
        \textbf{Gemini 2.0 Flash-Lite} & 0.976 & 0.389 & 0.364 \\\hline
        \textbf{Gemini 2.5 Flash} & 0.951 & 0.744 & \textbf{0.695}  \\\hline
        \textbf{GPT-4o Mini} & 0.908 & 0.511 & 0.419 \\ \hline

     \end{tabular}
    \caption{\change{Diagnostic capabilities of open-weight models over $\Mask$ in the \textit{anatomy} task (top). We also show the performance of commercial models (bottom).
    All models were evaluated with \PSNLF prompts.}} 
    \label{tab:correctnessopenweight} 
\end{table}

\change{
We tested the diagnostic capabilities of 
Mistral Small-2402, Llama 3-70b, Qwen3-1.7b,
and Qwen3-8b. Table \ref{tab:correctnessopenweight} shows the results for the \textit{anatomy} matching task using the \PSNLF prompts.
Qwen3-1.7b, the smallest model evaluated, performed as expected with limited success in diagnosing mappings within $\Mask$. In contrast, Qwen3-8b delivered highly competitive results, surpassing several larger commercial and open-weight models in terms of YI index. Appendix \ref{app:qwen} includes results for Qwen3 models for all prompt templates. Mistral Small-2402 also performed strongly, matching the level of Gemini 2.0 Flash and GPT-4o Mini. 
Despite its size, Llama 3-70b underperformed expectations.
%
%
These findings highlight the potential of open-weight models—whether accessed via APIs like Amazon Bedrock or deployed on local infrastructure—to serve effectively as \Oracles in ontology alignment tasks. Nonetheless, the choice of model involves balancing trade-offs among performance, latency, cost, and access to infrastructure.
Running open-weight models via Amazon Bedrock was generally more costly than commercial APIs, while local deployment significantly increased latency due to the limited resources.
}

\paragraph{Experiments on data leakage.}
Data leakage is a well-known challenge in the AI community, in general, and in the ontology matching community, in particular, when evaluating LLM-based systems on tasks with publicly available ground truths. We conducted an experiment using the OAEI NCIT–DOID dataset to assess the presence of potential critical data leakage. In the OAEI, ground truths are provided as sets of (correct) URI pairs (\eg equivalence relations between entities identified by their URIs, \ie their ontology ids). If the evaluated LLMs had been pretrained on these ground truths, a simple prompt such as ``Is URI1 equivalent to URI2?''---without any additional contextual information---would be expected to yield performance comparable to, or better than, the results reported in this study. However, this experiment resulted in a Youden's Index of approximately 0.01, indicating very limited evidence of leakage of the OAEI \emph{bio-ml} ground truths.


\section{Conclusions and future work}
\label{sec:conclusions}

The integration and understanding of the power of state-of-the-art LLMs within ontology alignment tasks is still at an early stage. Although the literature has shown promising results, there are still open challenges concerning performance, costs, and the sustainable use of LLMs. In this paper, we have explored the feasibility of integrating an LLM-based \Oracle with the state-of-the-art system \logmap, such that the \Oracle is only called for a very specific subset of mappings for which \logmap is uncertain. To the best of our knowledge, although LLMs are increasingly being used within ontology alignment pipelines, the use of LLMs as \Oracles has not been explored in the literature. We have provided an extensive evaluation of LLM-based Oracles as a diagnostic engine, as well as in combination with \logmap on an end-to-end ontology matching~task. The obtained results are encouraging, improving the performance of \logmap and achieving the top-2 results in the OAEI \emph{bio-ml} track. 
However, we have also shown that the results are far from a perfect \Oracle.

We foresee several promising directions for future work. One key avenue is to extend the contextual information of the prompts, leveraging additional ontological relationships. 
Exploring a broader range of prompt formulations and LLM models could also provide deeper insights into the opportunities of using LLMs as \Oracles.
Given the observed variation in performance across different prompts and models, combining multiple LLM-based \oracles through ensemble methods could result in more reliable outcomes and enhanced performance. 
%
%
Automatic prompt tuning and selection tailored to the matching task represents a promising direction for future work. We also plan to investigate few-shot prompts, particularly for tracks such as \textit{bio-ml}, where a subset of mappings can be leveraged for training.
%
Retrieval-augmented generation (RAG) may also enable systems to dynamically access relevant background knowledge, such as BioPortal \cite{bioportal2009}, leading to more informed and accurate diagnostic capabilities.

\section{Limitations}
\label{sec:limitations}

While our approach demonstrates strong results, several limitations merit discussion.


\paragraph{Missing human evaluation.} 
\change{A user study may also provide interesting insights in comparison with the LLM-based \Oracles. However, to make the exercise meaningful, we should involve domain experts in the process. In this paper, we have performed a comparison with simulated Oracles with different error rates, simulating a potential behaviour of a (non-perfect) domain expert. 
}

\paragraph{Potential training data leakage.}
Although our experiments on data leakage found no strong evidence of leakage of the OAEI ground truths, we cannot guarantee that the evaluated LLMs have not been 
exposed to existing OAEI benchmarks during pre-training, which could artificially boost their reported accuracy. To support a fair and unbiased evaluation of the new generation of ontology alignment systems relying on LLMs, the ontology matching community should prioritise the creation of new tasks with truly hidden (blind) reference alignments \change{as discussed during the ISWC 2024 special session on \emph{Harmonising Generative AI and Semantic Web Technologies: Opportunities, challenges, and benchmarks} \cite{alharbi2024llm}.
Nevertheless, the conducted experiments are still valid, even under the potential assumption of data leakage, as we are comparing the diagnostic capabilities of state-of-the-art models under the same conditions.}

\paragraph{Resource constraints.} Although our selected LLMs strike a balance between cost and quality, financial and infrastructure constraints still pose challenges for widespread adoption of LLM-based \Oracles, especially in large-scale or time-sensitive applications. Additionally, commercial model usage often involves rate limits and API changes, which could affect system stability.

\paragraph{Evaluation scope.} Our experiments focused on OAEI datasets within three tracks. While these cover a diverse set of biomedical domains and alignment challenges, additional evaluation on other OAEI datasets would be necessary to fully understand the robustness and limitations of our LLM-based \Oracle approach—\change{especially OAEI datasets in different domains and involving the matching of properties and instances}.

\paragraph{Focus on equivalence mappings.}
\change{\logmap can output both equivalence and subsumption mappings; indeed, it internally represents equivalence mapping as two subsumption mappings. Nevertheless, in this paper, we focus on the most common type of mapping: equivalence. In the future, LLMs can be leveraged to also discover and validate subsumption mappings and complex alignments (\ie arbitrary ontology axioms mentioning entities of two or more ontologies).
}

\paragraph{Evaluation with additional LLMs.} 
\change{Our evaluation relies on a subset of available LLM models selected according to limited financial and computational resources. We plan, however, to extend the evaluation with additional models that meet our resource limitations. We also intend to extend the evaluation to test open-weight models on the end-to-end alignment task.}

\section{Ethical consideration}
AI tools were used only for grammar and minor language edits. No content creation, idea development, or substantive rewriting was performed by AI. All research design, experiments, analysis, and writing were carried out by the authors.

\paragraph*{Contributions.} 
EJR and AG defined the research objectives. SL, DS, SS, and EJR jointly designed the methodology and experiments. SL and SS implemented the system infrastructure, including API integrations and formal evaluation procedures. SL led the ontology data acquisition, system optimization, and visualization components. DS designed and implemented the prompt-engineering framework and developed the infrastructure for integrating local language models.
EJR and AG contributed to the analysis of the experimental results. All authors contributed to drafting and revising the manuscript and approved the final version.

\paragraph*{Acknowledgement.}
This research was supported by the \href{https://r-ai.co/ukraine}{RAI for Ukraine} program of the NYU Center for Responsible AI, and by Turing Innovations Limited and The Alan Turing Institute’s Defence and Security Programme via the project \href{https://ernestojimenezruiz.github.io/projects/guard/}{GUARD}.
We would also like to acknowledge the support of Dr Dave Herron in the creation of an integrated pipeline for LogMapLLM, which enhances the reproducibility of the results.
Finally, we thank the anonymous reviewers for their valuable feedback during the ARR process.

\bibliographystyle{splncs04}
\bibliography{0references_arxiv_2026}

\appendix

\section{LogMap ontology alignment system}
\label{app:logmap}

\logmap is an ontology alignment system that combines lexical, structural, and semantic techniques with logical reasoning to efficiently match large ontologies while preserving consistency \cite{logmap2011,logmap2012}.\footnote{\url{https://github.com/ernestojimenezruiz/logmap-matcher}} It supports interactive alignment and has consistently performed well in the Ontology Alignment Evaluation Initiative (OAEI).

\begin{figure}[h!]
    \centering
    \includegraphics[width=0.80\linewidth]{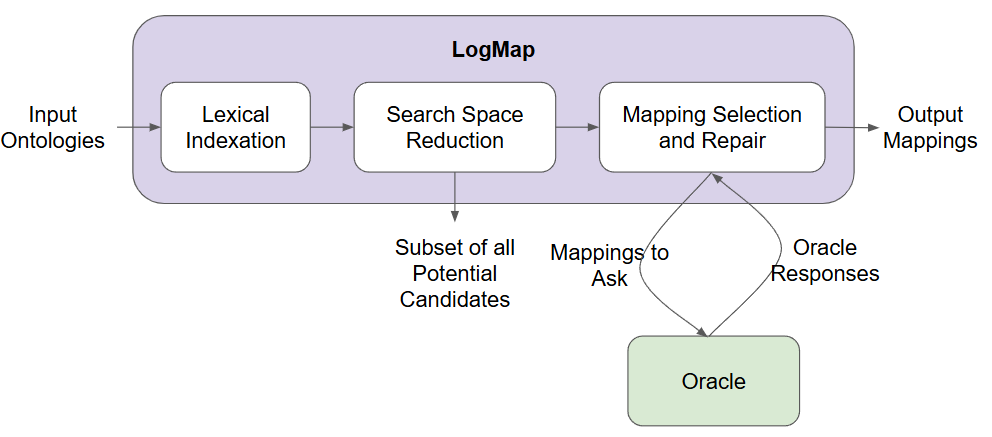}
    \caption{Workflow of the ontology alignment system \logmap with calls to an Oracle.}
    \label{fig:oa-system}
\end{figure}

\logmap defines some heuristics based on lexical and structural similarity to decide about the validity of a candidate mapping. However, some mappings are not clear-cut cases, and those are the mappings \logmap selects to ask an Oracle ($\Mask$).
Figure \ref{fig:oa-system} shows the workflow followed by \logmap when allowing interaction.
If an \oracle is not available, \logmap can also operate in an automatic (non-interactive) mode.

\section{Ontology-driven and system prompts}
\label{app:prompts}

Listings \ref{lst:pec}-\ref{lst:prompt_nlf_s_ec}
show the ontology-driven prompts with extended context. While Listing \ref{lst:systemprompt} shows the tested system prompts.


\begin{lstlisting}[basicstyle=\scriptsize\ttfamily,caption={\PEC prompt (non natural language-friendly with extended context).},%
                   label={lst:pec}]
Analyze the following entities, each originating from a distinct ontology. Each is represented by its **ontological lineage**, capturing its hierarchical placement from the most general to the most specific level.

1. Source entity ontological lineage:
	Level 0: alveolus epithelium
	Level 1: lung epithelium
	Level 2: respiratory system epithelium

2. Target entity ontological lineage:
	Level 0: Alveolar_Epithelium
	Level 1: Epithelium
	Level 2: Epithelial_Tissue, Normal_Tissue

Based on their **ontological positioning, hierarchical relationships, and semantic alignment**, do these entities represent the **same ontological concept**? Respond with "True" or "False".
\end{lstlisting}


\begin{lstlisting}[basicstyle=\scriptsize\ttfamily,caption={\PECNLF Prompt (natural-language friendly with extended context).},%
                   label={lst:prompt_ec_nlf}]
We have two entities from different ontologies.

The first one is "alveolus epithelium", which belongs to the broader category "lung epithelium", under the even broader category "respiratory system epithelium"

The second one is "Alveolar_Epithelium", which belongs to the broader category "Epithelium", under the even broader category "Epithelial_Tissue, Normal_Tissue"

Do they mean the same thing? Respond with "True" or "False".
\end{lstlisting}



\begin{lstlisting}[basicstyle=\scriptsize\ttfamily,caption={\PECSNLF Prompt (natural-language friendly with synonyms and extended context).},%
                   label={lst:prompt_nlf_s_ec}]
We have two entities from different ontologies.

The first one is "alveolus epithelium", belongs to broader category "lung epithelium", under the even broader category "respiratory system epithelium" (also known as "respiratory system mucosa").

The second one is "Alveolar_Epithelium", also known as "Alveolar Epithelium", "Lung Alveolar Epithelia", "Epithelia of lung alveoli", belongs to broader category "Epithelium" (also known as "Epithelium", "epithelium"), under the even broader category "Epithelial_Tissue, Normal_Tissue".

Do they mean the same thing? Respond with "True" or "False".
\end{lstlisting}


\begin{lstlisting}[basicstyle=\scriptsize\ttfamily,caption={System prompts},%
                   label={lst:systemprompt},
                   breaklines=true,
                   breakatwhitespace=true,
                   linewidth=\linewidth]
Base = "You are a professional ontology matcher. You need to answer different questions about matching ontologies. Be precise."

Explainable = "You are helping researchers determine if two biomedical terms from different ontologies refer to the same concept. You'll be provided with a natural-language description, possibly including synonyms and parent categories. Think like a domain expert, but explain your judgment intuitively. Be precise"

Hierarchical = "You are a biomedical ontology expert. Your task is to assess whether two given entities from different biomedical ontologies refer to the same underlying concept. Consider both their semantic meaning and hierarchical context, including parent categories and ontological lineage. Be precise."

Lexical = "You are a domain expert assisting in entity alignment across biomedical ontologies. Each entity may include synonyms and category-level relationships. Use synonym information and parent class semantics to decide whether the two entities mean the same thing. Be precise."

\end{lstlisting}

\section{Additional supporting results}
\label{app:results}

\begin{figure*}[t!]
     \centering
     \includegraphics[width=0.999\linewidth]{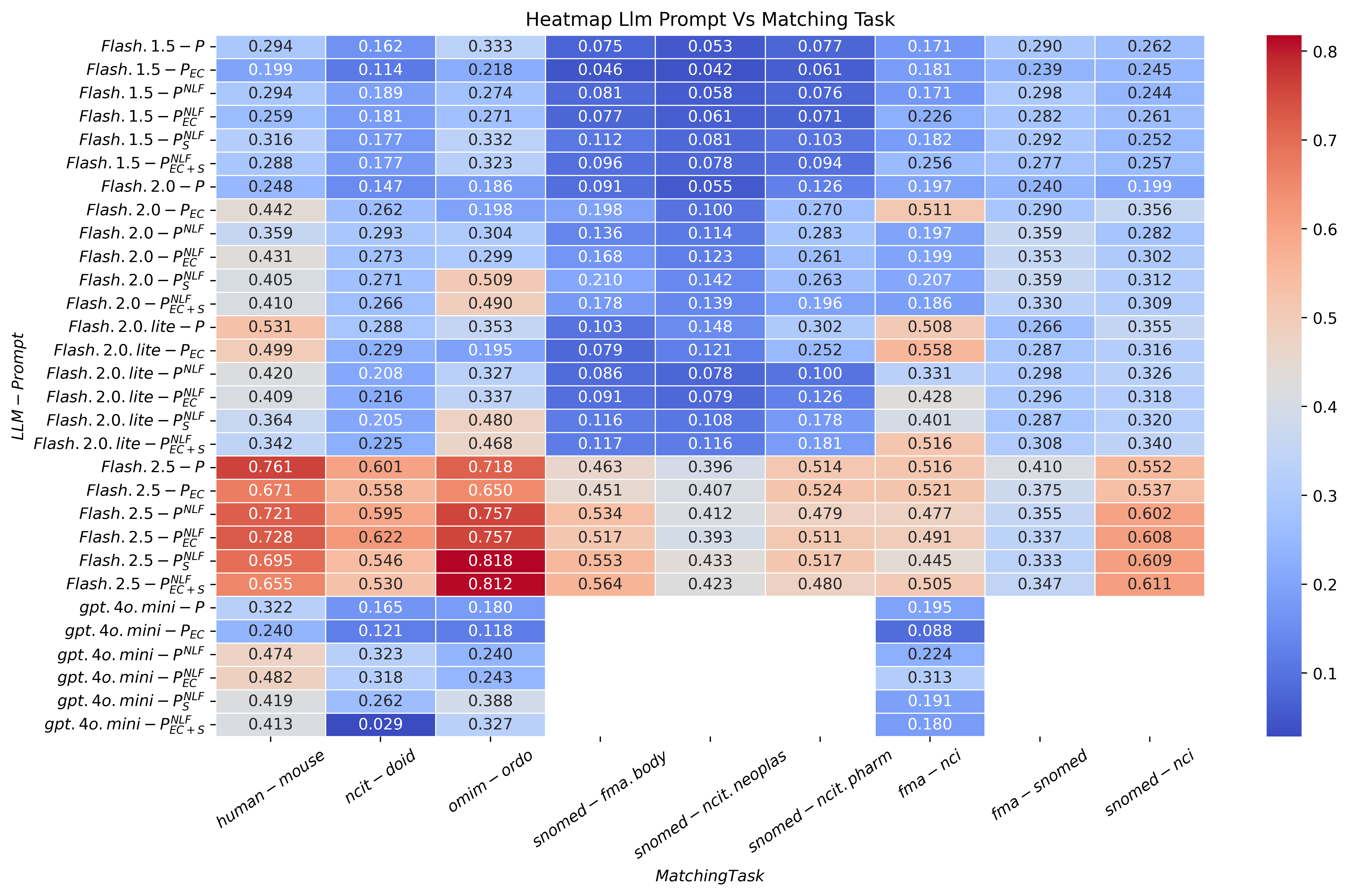}
     \caption{Diagnostic results (Youden’s index) by the LLM-based \oracles over the selected ontology matching tasks. For example, \emph{Flash 2.5-\PECSNLF} represents the LLM-based \Oracle relying on the Gemini Flash 2.5 model and evaluated with the natural-language friendly (NLF) prompts with extended context (EC) and synonyms (S). We only completed a subset of experiments with GPT-4o Mini as a reference.}
     \label{fig:all-results}
 \end{figure*}

Figure \ref{fig:all-results} shows the 
correctness (Youden’s index, YI) of the LLM-based \oracles in assessing the mappings in $\Mask$ (\ie the subset of mappings identified by \logmap as uncertain).
We tested, over the 9 matching tasks, a total of 30 LLM-based \Oracles (\OrLLM), combining the six prompt templates introduced in Section \ref{sec:templates} and the LLM models presented in Section \ref{sec:llm}.

\begin{figure*}[t!]
    \centering
    \includegraphics[width=0.999\linewidth]{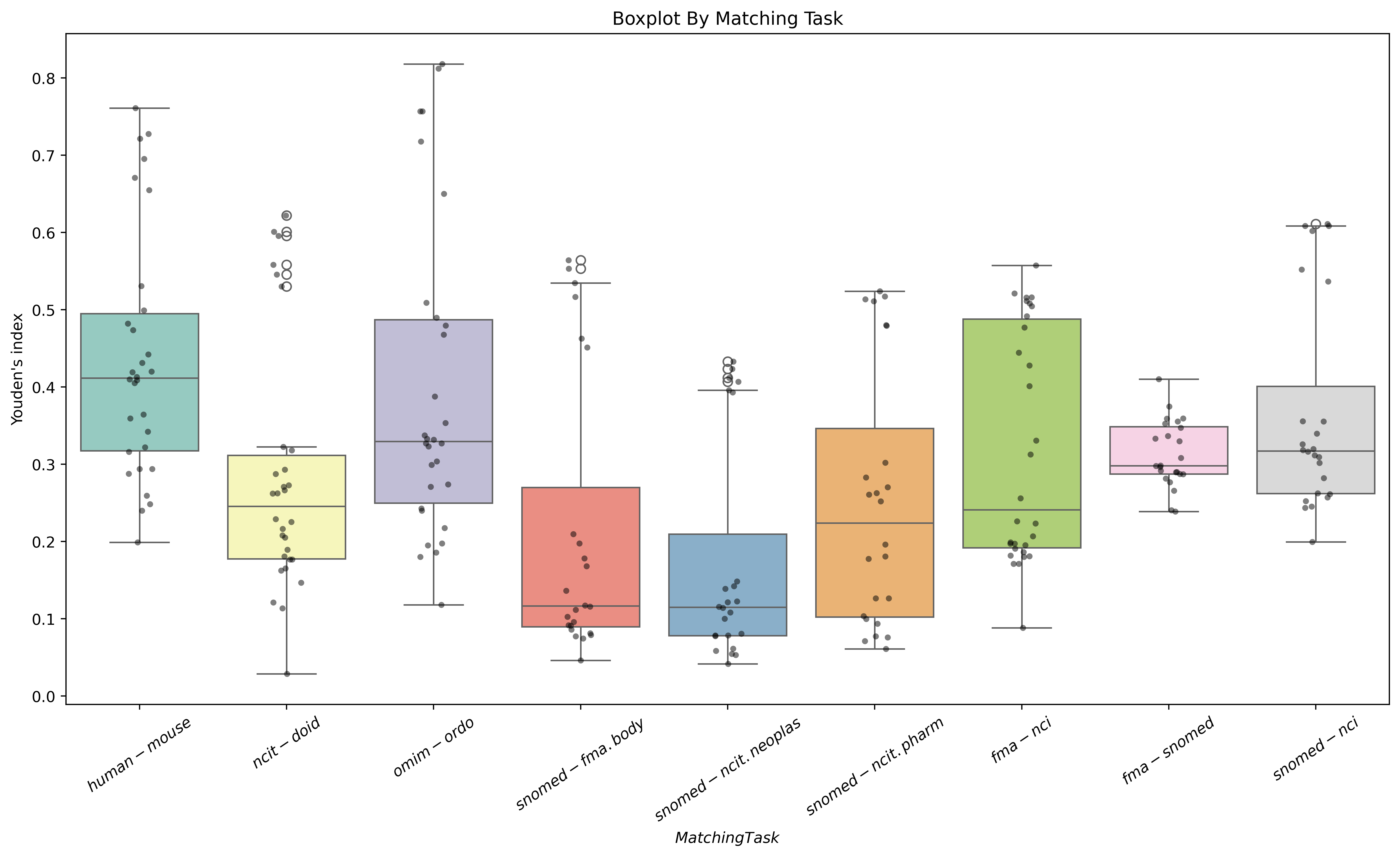}
    \caption{Average YI values per ontology matching task, reflecting the varying complexity of $\Mask$ across tasks.}
    \label{fig:by_task}
\end{figure*}

Figure \ref{fig:by_task} shows the average YI values across ontology matching tasks, highlighting the differing levels of complexity in $\Mask$ for each task.

\section{Experiments on determinism}
\label{app:det}

\begin{figure}[h!]
    \centering
    \includegraphics[width=0.9\linewidth]
    {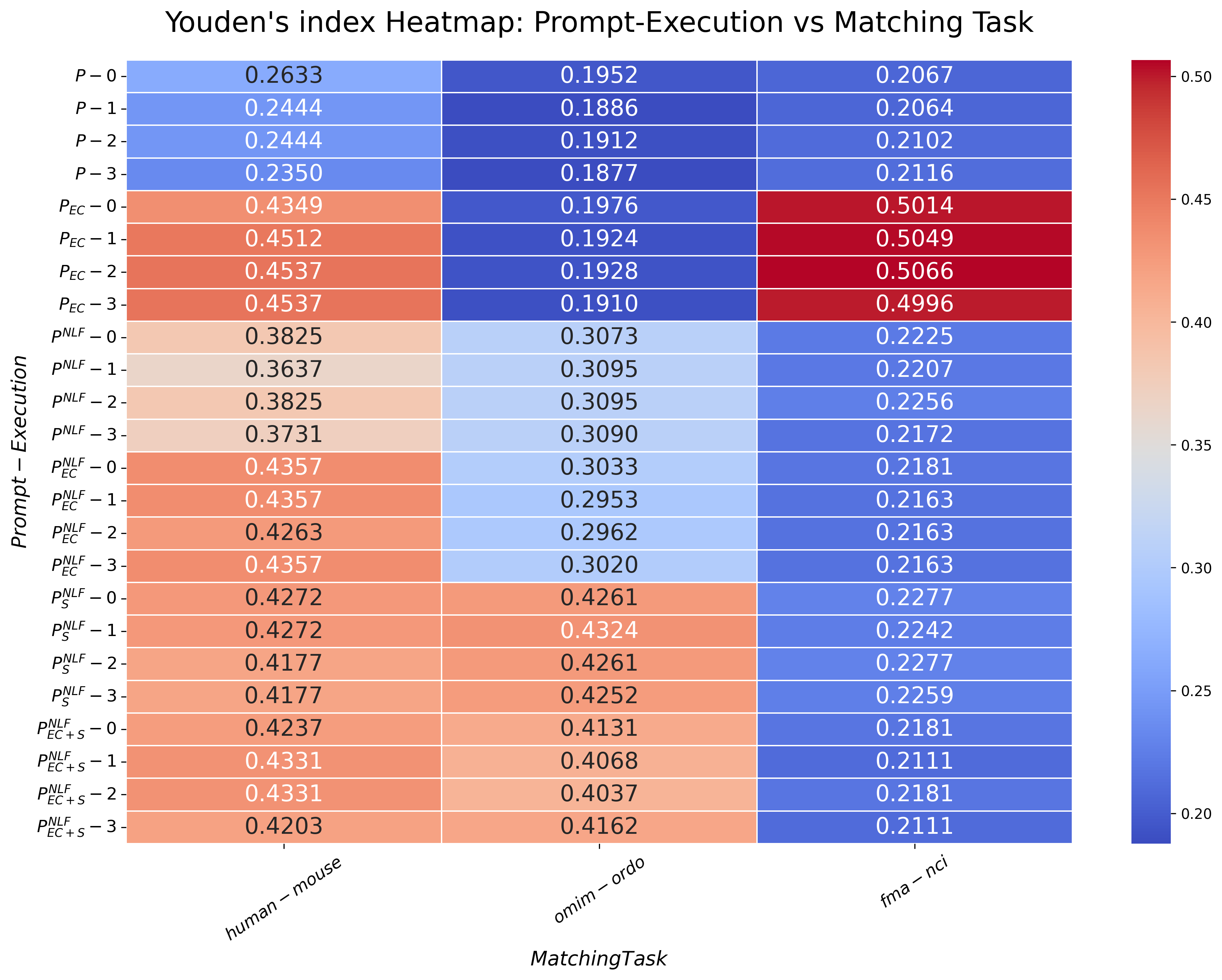}
    \caption{Determinism of Gemini 2.0 Flash across four runs for three matching tasks and all prompt templates.}
    \label{fig:determinism}
\end{figure}

 Figure \ref{fig:determinism} shows the variations of the YI index across 4 independent runs with all six prompt templates on three ontology matching tasks using Gemini Flash 2.0.

\begin{figure}[h!]
    \centering
    \includegraphics[width=0.70\linewidth]
    {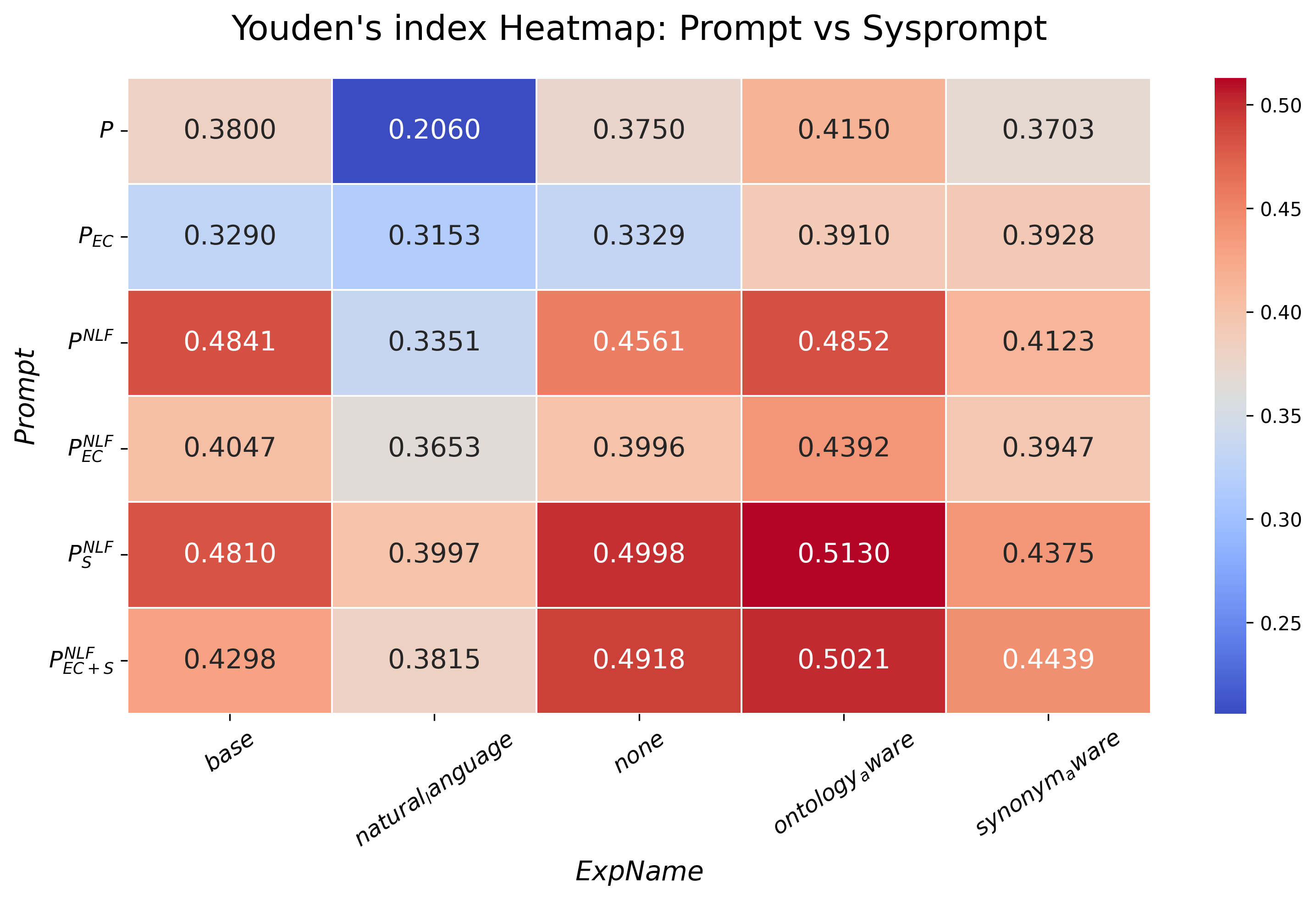}
    \caption{Performance variation of Gemini 2.0 Flash with different system prompts.}
    \label{fig:determinismSP}
\end{figure}

Figure \ref{fig:determinismSP} illustrates the effect of different system prompts on the average YI index, revealing 
performance differences across prompts.

\section{Experiments on statistical analysis}
\label{app:stat}

We conducted t-test and Wilcoxon statistical tests to analyse whether the performance differences reported in Table \ref{tab:fulltask} and Figure~\ref{fig:radar-error-rate} were significant ($p$-value $< 0.05$). 
Table~\ref{tab:stat_tests} confirms that \logmapP\OrZero and \logmapP\OrTen lead to significantly better results than both \logmapP\OrGem and \logmapP\OrGemNew (\ie $p < 0.01$ in the ``less'' and ``two-sided'' settings with both t-test and Wilcoxon).
The comparison of \logmapP\OrGem and \logmapP\OrGemNew with \logmapP\OrTwenty yields no significant differences ($p > 0.1$ in the two-sided setting). 
\logmapP\OrGem and \logmapP\OrGemNew also significantly improve \logmapP\OrThirty and \logmap in automatic mode (\ie $p < 0.05$ in the ‘greater’ direction for both tests).
Overall, the statistical tests support the results and the discussion presented in Section \ref{sec:eval}.

\setlength{\tabcolsep}{0.005pt}
\begin{table}[t]
    \centering
    \begin{tabular}{|l||c|c|c||c|c|c|}
        \hline
        \multirow{2}{*}{\textbf{System Comparison}} & \multicolumn{3}{c||}{\textbf{t\_test}} & 
        \multicolumn{3}{c|}{\textbf{Wilcoxon}}  \\ \cline{2-7}  
        & \textbf{greater} & \textbf{less} & \textbf{2-sided} & 
          \textbf{greater} & \textbf{less} & \textbf{2-sided}  \\\hline\hline
        \textbf{\logmapP\OrGem vs \logmapP\OrZero} & ~0.999~ & ~0.0001~ & ~0.0002~ & 1.0 & 0.002 & ~0.004~ \\ \hline
        \textbf{\logmapP\OrGem vs \logmapP\OrTen} & 0.999 & 0.0003 & 0.0006 & 1.0 & 0.002 & 0.004 \\ \hline
        \textbf{\logmapP\OrGem vs \logmapP\OrTwenty} & 0.918 & 0.082 & 0.165 & 0.898 & 0.125 & 0.25 \\ \hline
        \textbf{\logmapP\OrGem vs \logmapP\OrThirty} & 0.030 & 0.970 & 0.060 & ~0.037~ & ~~0.973~~ & ~0.074~ \\ \hline
        \textbf{\logmapP\OrGem vs \logmap} & 0.0007 & 0.999 & 0.001 & 0.002 & 1.0 & 0.004 \\ \hline\hline

        \textbf{\logmapP\OrGemNew vs \logmapP\OrZero} & 0.998 & 0.002 & 0.003 & 1.0 & 0.002 & 0.004 \\ \hline
        \textbf{\logmapP\OrGemNew vs \logmapP\OrTen} & 0.991 & 0.009 & 0.018 & 0.998 & 0.004 & 0.008 \\ \hline
        \textbf{\logmapP\OrGemNew vs \logmapP\OrTwenty} & 0.797 & 0.203 & 0.406 & 0.787 & 0.248 & 0.496 \\ \hline
        \textbf{\logmapP\OrGemNew vs \logmapP\OrThirty} & 0.037 & 0.963 & 0.074 & 0.049 & 0.963 & 0.098\\ \hline
        \textbf{\logmapP\OrGemNew vs \logmap} & 0.020 & 0.980 & 0.041 & 0.027 & 0.981 & 0.055 \\ \hline

    \end{tabular}
    \caption{Statistical test results comparing \logmap, \logmap with \OrGem and \OrGemNew, and \logmap in combination with Oracles with different error rates (\OrZero, \OrTwenty, and \OrThirty). Values represent $p$-values for t-test and Wilcoxon signed-rank test in \emph{greater}, \emph{less}, and \emph{two-sided} settings.}
    \label{tab:stat_tests}
\end{table}

\section{Additional experiments with Qwen models}
\label{app:qwen}

Table \ref{tab:correctnessqwen} compares the results of Qwen3-1.7b and Qwen3-8b  across the different prompt templates. Qwen3-1.7b produces very low scores, typically diagnosing most mappings as negative. Qwen3-8b performed well with the natural-language-friendly prompt templates. Structured prompts, however, led to poor diagnostic capabilities, similar to Qwen3-1.7b.

\setlength{\tabcolsep}{3.25pt}
\begin{table}[tb!]
    \centering
    \begin{tabular}{|l||c|c|c||c|c|c|}
        \hline
        \multirow{2}{*}{\textbf{Prompt}} & \multicolumn{3}{c||}{\textbf{Qwen3-1.7b on $\Mask$}} & \multicolumn{3}{c|}{\textbf{Qwen3-8b on $\Mask$}}   \\ \cline{2-7}  
        &
        \textbf{~~~Se~~~} & \textbf{~~~Sp~~~} & \textbf{~~~YI~~~} & \textbf{~~~Se~~~} & \textbf{~~~Sp~~~} & \textbf{~~~YI~~~} \\ \hline\hline
         \textbf{\Prompt} & 0.096 & 0.972 & 0.068  
         & 0.082 & 1.000 & 0.082\\\hline
         \textbf{\PEC}  & 0.058 & 0.981 & 0.039 
         & 0.055 & 0.991 & 0.045\\\hline
         \textbf{\PNLF}  & 0.199 & 0.925 & 0.123
         & 0.483 & 0.943 & 0.426 \\\hline
         \textbf{\PECNLF}  & 0.168 & 0.962 & 0.130
         & 0.435 & 0.962 & 0.397\\\hline
         \textbf{\PSNLF}  & 0.411 & 0.811 & 0.222
         & 0.825 & 0.764 & \textbf{0.590}\\\hline
        \textbf{\PECSNLF} & 0,414 & 0.906 & 0.320
         & 0.688 & 0.793 & 0.481 \\\hline
     \end{tabular}
    \caption{Performance of evaluated Qwen models in the \textit{anatomy} task across the six prompt templates. Se=Sensitivity, Sp=Specificity, YI=Youden’s index. NLF=Natural-language friendly, EC=Extended context, S=Synonyms.}       
    \label{tab:correctnessqwen} 
\end{table}

\end{document}